\documentclass{article}
\usepackage{spconf,amsmath,epsfig}

\usepackage{cite}

\usepackage{graphicx}
\graphicspath{{Figures/}}
\DeclareGraphicsExtensions{.pdf,.jpeg,.png}

\usepackage{amsmath}
\usepackage{algorithmic}

\usepackage{gensymb}
\usepackage{subfloat}

\usepackage{array}

\usepackage{multirow}
\usepackage{amsfonts}
\usepackage{stfloats}

\usepackage{url}
\usepackage[export]{adjustbox}
\usepackage{textcomp}
\usepackage{cite}

\usepackage{caption}
\usepackage[font=footnotesize]{subcaption}
\usepackage{breqn}
\usepackage{lipsum}
\usepackage{framed,multirow}
\usepackage{booktabs}
\usepackage{url}
\usepackage{xcolor}

\usepackage{comment}

\usepackage{algorithm}
\usepackage{eqparbox}

\usepackage{graphicx}
\usepackage{amsmath,amssymb} %
\usepackage{bm}
\usepackage{booktabs}
\usepackage{rotating,multirow}
\usepackage{wrapfig}

\usepackage{tikz}
\usepackage[export]{adjustbox}
\usepackage[inkscapearea=page]{svg}
\usepackage{svg-extract}

\graphicspath{{./Images/}{./Figures/result/}{./Images/result/}{./svg-inkscape/}}

\begin{document}

	\title{To Be Defined: Knowledge Distillation and Object Detection}
    \title{Knowledge Distillation for Object Detection: \\ from generic 
    to remote sensing datasets}
    \name{Hoàng-Ân Lê and Minh-Tan Pham~\thanks{This work was supported by the SAD 2021 ROMMEO project (ID 21007759) and the ANR AI chair OTTOPIA project (ANR-20-CHIA-0030).}}
    \address{IRISA, Université Bretagne Sud, UMR 6074, 56000 Vannes, France \\
    \tt\small \{hoang-an.le,minh-tan.pham\}@irisa.fr}
	\maketitle
	
\begin{abstract}
    
Knowledge distillation, a well-known model compression technique,
is an active research area in both computer vision and
remote sensing communities.
In this paper, we evaluate in a remote sensing context
various off-the-shelf object detection knowledge
distillation methods which have been originally developed on generic computer
vision datasets such as Pascal VOC. In particular,
methods covering both logit mimicking and feature
imitation approaches are applied for vehicle detection
using the well-known benchmarks such as xView and VEDAI datasets.
Extensive experiments are performed to compare the
relative performance and interrelationships of the methods.
Experimental results show high variations 
and confirm the importance of result aggregation
and cross validation on remote sensing datasets.

\end{abstract}


\section{Introduction}
\label{sec:intro}

Deep learning has proven to be a potential method for inter-disciplinary image
analysis and scene understanding. However, the well-known data-demanding nature 
and expensive computation tendency limit its applicability on  resource-constrained edge
devices, such as cellphones in the ordinary context or
satellites and unmanned aerial vehicles, in remote sensing (RS).
Model compression methods aim to simplify a deep model without diminishing too much performance,
among which knowledge distillation (KD) techniques are prominent.
Generalized by Hinton~\textit{et al.}~\cite{Hinton2015} for deep convolutional neural networks,
KD trains a small model, called
\textit{student} to follow a distribution produced by a large and cumbersome
model, called \textit{teacher} and has been an active research area of the
machine learning and computer vision community since.
Several methods have been proposed first for image classification
and extended to other tasks such as object detection and
semantic segmentation ~\cite{Hinton2015,Zhang2021PDF,Hou2020Road}.

Efforts have been made to apply KD in RS contexts.
As examples, Liu~\textit{et al.}~\cite{Liu2021Zoom} follows the image
enlargement idea to enhance the features of small objects by introducing a
cross-scale KD method for drone-based object detection;
Li~\textit{et al.}~\cite{Li2021KP} reduces noise and maintains semantic information
in a keypoint-based detection pipeline using a semantic transfer block to merge
shallow and deep features before distilling information to a student network,~\textit{etc}.
Despite promising achievements, contributions of such work mostly involve %
designing mechanisms to cope with a particular
RS challenge, such as small or varied object sizes~\cite{Liu2021Zoom}, %
arbitrary orientations and complex backgrounds~\cite{Li2021KP},~\textit{etc.},
while the principal structures and setups are often inherited from common vision practices,
which might not always be suitable due to inherent differences
between RS and generic datasets with which they were originally developed.

In this work, we evaluate several off-the-shelf KD
methods for object detection and showcase their relative performances
on generic and RS datasets. The purpose is to present to the RS
community the performance transition %
of the same methods from the originally developing domain to RS
so that correspondences can be made and possible insights could be drawn
for future developments.
To that end, we revisit the original setup and the core ideas of KD for object detection,
including both logit mimicking and feature imitation~\cite{Zheng2022LD}.
Training a student network using the teacher's
probability predictions as \textit{soft} targets, i.e. logit mimicking,
in place of \textit{hard}, one-hot vector labels
show benefits as they contain richer information such as the relationship between
classes~\cite{Hinton2015}. However, as logit mimicking is known to be ineffective for
object detection~\cite{Kang2021InstCond,Wang2019finegrain,Zhang2021FeatKD} as it
can only transfer semantic knowledge and neglects localization information,
feature imitation methods enforcing the consistency of deep features between teachers
and students.

\section{Method}
\label{sec:method}

\begin{figure*}[t]
    \centering
    \scriptsize
    \def\svgwidth{.95\textwidth}
    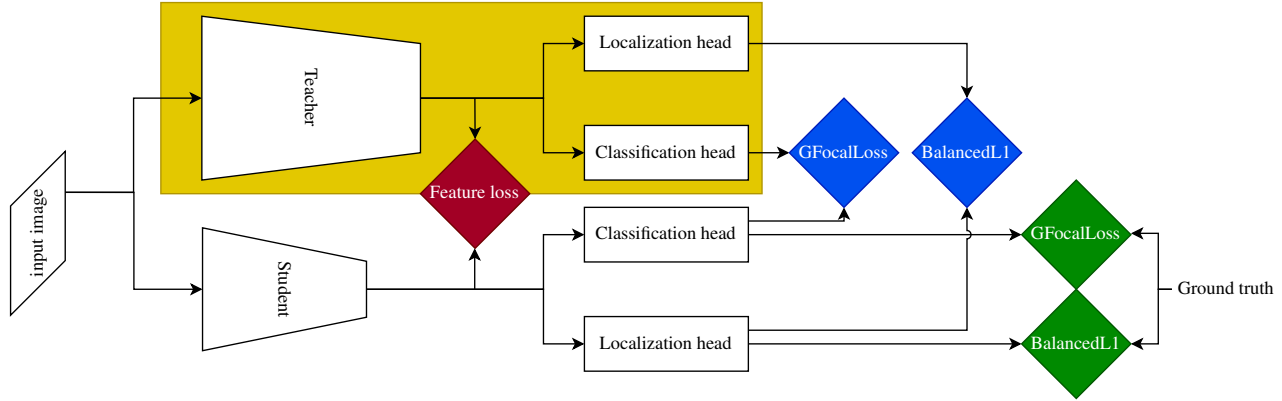
    \caption{Knowledge distillation pipeline: the teacher branch (in yellow) is trained once and set frozen when training the student network. Either hard- (green squares) or soft-target (blue) loss is needed to train the classification head. Feature loss  (red) is to enforce feature imitation.
    }
    \label{fig:illustration}
\end{figure*}

Knowledge distillation involves utilizing similar networks of different capacities,
a larger and more cumbersome teacher and smaller student, so that the smaller network
performance can be improved by training and distilling information from the larger one.
In this work, the well-known RetinaNet architecture~\cite{Lin2017focal} for
one-stage object detection is employed due to their simplicity.
An illustration is shown in Figure~\ref{fig:illustration}.

\subsubsection*{Object Detection}

The encoder part includes a backbone from, chosen from the ResNet family,
responsible for extracting features from input images, and a FPN-family neck
to aggregate extracted features at multiple scales.
For the teacher network, the ResNet50 backbone with PAFPN neck~\cite{Liu2018PAFPN} are used
and for the student, ResNet18 and FPN neck~\cite{Lin2017}. The teacher-student parameter ratio is
1.6. The aggregated features from the neck are passed into the detection head, comprising a
classification and localization subnets.

A few modifications are made following the implementation of~\cite{Zhang2021PDF}, including
removing the first max-pooling layer of ResNet as in ScratchDet~\cite{Zhu2019scratchdet},
and adding the context enhancement module as in ThunderNet~\cite{Qin2019thundernet}.
The number of convolutional blocks in each subset is reduced from 4 to 2 to speed
up the training time. The teacher network is set frozen during
the training of the student.

The Balanced L1 Loss~\cite{Pang2019} is employed for localization loss,
defined on the L1-difference between predicted and target boxes.

For classification, the multiple binary classification with
sigmoid operation is adopted for multi-class implementation,
following~\cite{Li2020gfocal,Lin2017focal,Tian2021fcos}.
The Quality Focal Loss~\cite{Li2020gfocal} joining the localization quality and classification score
is employed with soft-target given by the
IOU scores of predicted localization and ground truth boxes.

\subsubsection*{Knowledge Distillation}

Logit mimicking distillation trains a student network to follow the prediction made by a teacher
by using the teacher's probability predictions as soft targets while
feature imitation enforces the similarity at the feature level. As feature imitation does not
involve the detection heads, either hard (ground truth) or soft labels, i.e.
teacher predictions, have to be employed to train the student network successfully.

Three feature-imitation strategies are employed, namely mean-squared-error
difference (MSE), PDF-Distil (coined PDF) ~\cite{Zhang2021PDF}, and Decoupled
Features (coined DeFeat)~\cite{Guo2021Defeat}.
MSE minimizes the differences between the teacher's and student's across the feature maps equally
while PDF-Distil employs the teacher-student prediction disagreements to
dynamically generate a weighting mask for distillation and
DeFeat distills features separately from the foreground and background regions
with different assigned importance.
As the student's feature maps have different shapes
than the teacher's, an adaptive layer with $3\times3$ convolution is employed
before applying the loss function.

\begin{figure*}
    \centering \scriptsize
    \def\svgwidth{\textwidth}
    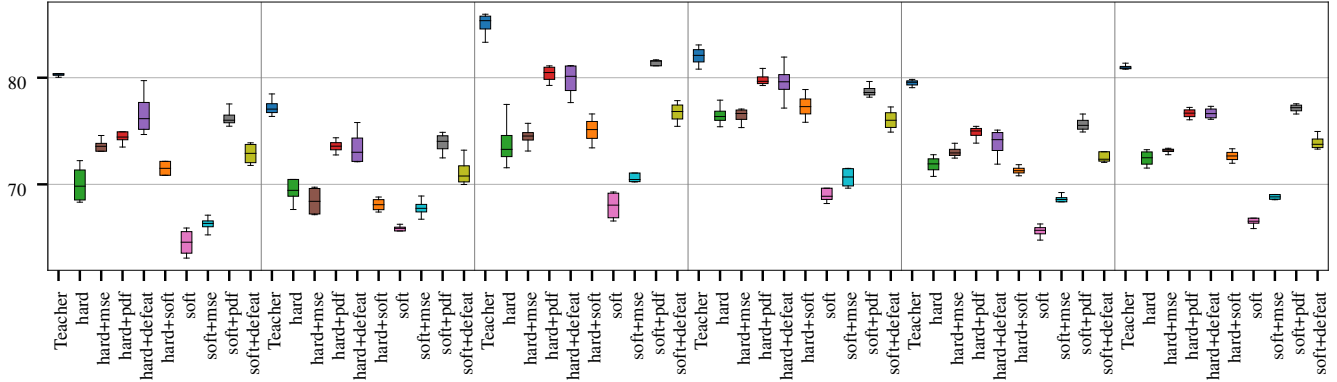
    \caption{
    MAP performance distribution with different random seeds on
    VEDAI25 for the first 5 official cross-validation folds (first 5 columns) and
    average (last column). Whisker value is set to 1.5 of the interquartile range of the MAPs.
    The performance varies greatly with the random seeds and across
    the validation fold. The average across 5 folds show stability.
    } 
    \label{fig:vedai25_distribution}
\end{figure*}

\section{Experiments}
\label{sec:exp}

For experimental setup, we use the Pascal VOC dataset~\cite{PascalVOC}, a
well-known benchmark in computer vision for the generic dataset, and 
two remote sensing datasets, following~\cite{Le2022RS}, namely xView~\cite{xView} and
VEDAI~\cite{vedai} which represents the satellite and aerial imagery with various
resolution levels. The xView data resolution is of 30cm while VEDAI data are available
in both 25-cm and 12.5-cm resolutions, termed VEDAI25 and VEDAI12 in this work.

We experiment with various distillation configurations, including
(1) hard-label and soft-label losses,
(2) hard-label and feature losses,
(3) only soft-label loss, and
(4) soft-label and feature losses.
The configuration using only hard labels is used as the baseline.

\subsubsection*{Performance distribution on VEDAI25}

As it is well-known that object detection performance on VOC benchmark is somewhat stable,
we first show the performance distribution of detection networks on the VEDAI25 dataset.
To that end, the first 5 official cross-validation folds are used.
For each configuration, we run the same network 4 times
with different random factors and show the MAPs distribution with a boxplot
in Fig.~\ref{fig:vedai25_distribution}.
The whisker value is set to 1.5 of the interquartile which is the range between
the first and third quartile of the MAPs.

It can be seen that the detection performance varies greatly for each
configurations in each of the 5 cross-validation folds (the first 5 columns)
and across the folds. The teachers' performances seem to be more stable than the
students' in each validation fold, and mostly stay above 80\%MAP mark across the folds.
The hard and hard+defeat configurations shows large random variations while hard+pdf
seems to be more condensed. Consequently, the individual inter-relationships among
the KD configurations also vary, considering the range of each configuration in
and across the folds. However, the average across 5-fold validations
(the last column) show the concentration of the performance
and thus the inter-relationship are more pronounced and conclusive, hinting
the good practice of using multiple cross validation sets when benchmarking on the
VEDAI dataset.

\subsubsection*{Knowledge distillation performance}

In Table~\ref{tab:hard_kd} and Table~\ref{tab:soft_feat}, the performances on 
all 3 datasets are shown numerically. For VEDAI25, the median of 4 runs is
shown using the 5-fold average (the last column in Fig.~\ref{fig:vedai25_distribution}).

Table~\ref{tab:hard_kd} shows the improvement when adding a
distillation method to a regular \textit{hard} training using one-hot
ground truth vectors for student networks.
The interrelationships between teacher-student and among the students for 
VEDAI25 are similar to VOC's and different for the other 2 datasets
as they are from 1-time runs (with the single 1st cross-validation fold is used
for VEDAI12) while VEDAI25 results are medians of multiple runs
and averaged across 5 cross-validation folds.

For the VOC dataset and VEDAI25, the performance increments from feature imitation
(MSE, PDF, DeFeat) are more than from logit mimicking while the relationships are
reversed for the other two datasets: using soft-target loss
outperforms MSE and DeFeat and on par with PDF.
This seems counter-intuitive since with smaller bounding boxes from
heterogeneous backgrounds, it might
be that RS data require more accurate localization, hence
should receive more benefit from feature imitation.
    
\begin{table*}[t]
    \begin{minipage}[t]{0.48\linewidth}
        \centering
        \setlength{\tabcolsep}{3pt}
        \begin{tabular}{@{}lcccc@{}}
            \toprule
            KD          &  VOC  & xView & VEDAI25& VEDAI12 \\ %
            \midrule                                             
            teacher     & 55.67 & 84.02 & 80.91  & 85.46   \\ %
            hard        & 50.12 & 76.06 & 72.49  & 76.57   \\ %
            \midrule                                             
            hard+soft   & 51.22 & 77.27 & 72.66  & 80.51   \\ %
            hard+MSE    & 52.29 & 76.84 & 73.22  & 77.75   \\ %
            hard+PDF    & 52.94 & 77.99 & 76.68  & 81.10   \\ %
            hard+DeFeat & 53.91 & 76.73 & 76.63  & 78.36   \\ %
            \bottomrule
        \end{tabular}
        \caption{Comparison of different types of distillation.
        Adding knowledge distillation to hard-label improves performance in general.
        PDF distillation has the best performance for the remote sensing datasets while it is DeFeat
        for VOC.
        }
        \label{tab:hard_kd}
    \end{minipage}
    \hfill
    \begin{minipage}[t]{0.47\linewidth}
        \centering
        \setlength{\tabcolsep}{3pt}
        \begin{tabular}{@{}lcccc@{}}
            \toprule
            KD          & VOC   & xView & VEDAI25 & VEDAI12\\ %
            \midrule                                             
            teacher     & 55.67 & 84.02 & 80.91   & 85.46  \\ %
            hard        & 50.12 & 76.06 & 72.49   & 76.57  \\ %
            \midrule                                             
            soft        & 50.44 & 70.23 & 66.54  & 78.15   \\ %
            soft+MSE    & 50.48 & 71.15 & 68.83  & 78.44   \\ %
            soft+PDF    & 52.24 & 72.21 & 77.19  & 80.13   \\ %
            soft+DeFeat & 51.16 & 74.07 & 73.75  & 79.21   \\ %
            \bottomrule
        \end{tabular}
        \caption{Using only soft labels diminishes the results hard labels for
    xView and VEDAI25 datasets but reverses for VEDAI12 and on par for VOC.
    Combining with feature distillation helps improving performance yet cannot
    reach the hard-labels performance for xView.
    }
        \label{tab:soft_feat}
    \end{minipage}
\end{table*}

Table~\ref{tab:soft_feat} shows the performance of the combination of distillation methods.
Similar to xView, the results using soft labels on VEDAI25 plunge in contrast to 
VEDAI12 and VOC. Adding MSE feature imitation slightly helps for
all datasets. Both PDF and Defeat distillation show further improvement where PDF performance
is the best on all datasets except xView, showing that for xView, 
the background information is more important than inter-class relationship, as there is only
1 class for prediction.

\section{Conclusion}
\label{sec:conclusion}

The paper evaluates various knowledge distillation methods
on both generic and remote sensing datasets and presents
the interrelationship and performance differences for each case.
While the performances are generally stable generic datasets such as
VOC, results on VEDAI are highly varying,
and thus, unless aggregated over multiple random-initialized runs with
different cross-validation sets, the results appear to be
random and inconclusive.

{\small
\bibliographystyle{IEEEbib}
\bibliography{1_main}
}

\end{document}